\documentclass[11pt,a4paper]{article}
\usepackage[hyperref]{acl2020}
\usepackage{times}
\usepackage{latexsym}

\usepackage{amssymb}
\usepackage{amsmath}
\usepackage{multirow}
\usepackage{multicol}
\usepackage{bm}
\usepackage{color}
\usepackage{graphicx}
\usepackage{url}
\usepackage{subfigure}

\usepackage{microtype}

\aclfinalcopy 

\title{Curriculum Pre-training for End-to-End Speech Translation}
\author{ \textbf{Chengyi Wang\thanks{\,\,Works are done during internship at Microsoft} \textsuperscript{\rm 1}, Yu Wu\textsuperscript{\rm 2}, Shujie Liu\textsuperscript{\rm 2}, Ming Zhou\textsuperscript{\rm 2}, Zhenglu Yang\textsuperscript{\rm 1}}\\ 
\textsuperscript{\rm 1}Nankai University, Tianjin, China\\
\textsuperscript{\rm 2}Microsoft Research Asia, Beijing, China\\
cywang@mail.nankai.edu.cn,  Wu.Yu@microsoft.com,  \\
shujliu@microsoft.com, mingzhou@microsoft.com, yangzl@nankai.edu.cn
\\ 
}

\date{}

\begin{document}
\maketitle
\begin{abstract}
End-to-end speech translation poses a heavy burden on the encoder, because it has to transcribe, understand, and learn cross-lingual semantics simultaneously. To obtain a powerful encoder, traditional methods pre-train it on ASR data to capture speech features. However, we argue that pre-training the encoder only through simple speech recognition is not enough and high-level linguistic knowledge should be considered. Inspired by this, we propose a curriculum pre-training method that includes an elementary course for transcription learning and two advanced courses for understanding the utterance and mapping words in two languages. The difficulty of these courses is gradually increasing. Experiments show that our curriculum pre-training method leads to significant improvements on En-De and En-Fr speech translation benchmarks.
\end{abstract}

\section{Introduction}
Speech-to-Text translation (ST) is essential to breaking the language barrier for communication. It aims to translate a segment of source language speech to the target language text.  To perform this task, prior works either employ a cascaded method where an automatic speech recognition (ASR) model and a machine translation (MT) model are chained together or an end-to-end approach where a single model converts the source language audio sequence to the target language text sequence directly \cite{DBLP:journals/corr/BerardPSB16}.

\begin{figure}[t]
    \centering
    \begin{subfigure}[previous encoder pre-training]{
    \begin{minipage}[t]{0.42\textwidth}
    \includegraphics[width=\textwidth]{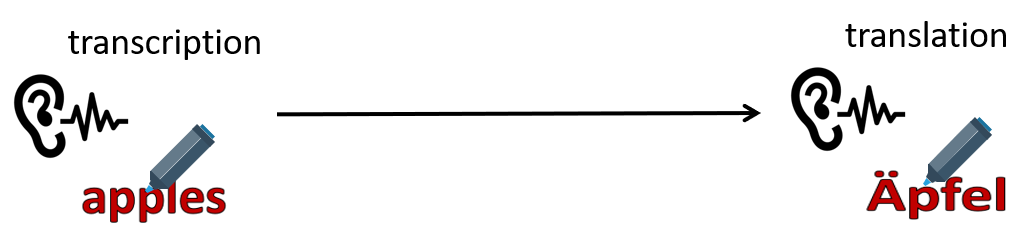}
    \end{minipage}
    }
    \end{subfigure}

    \begin{subfigure}[curriculum encoder pre-training]{
    \begin{minipage}[t]{0.42\textwidth}
    \includegraphics[width=\textwidth]{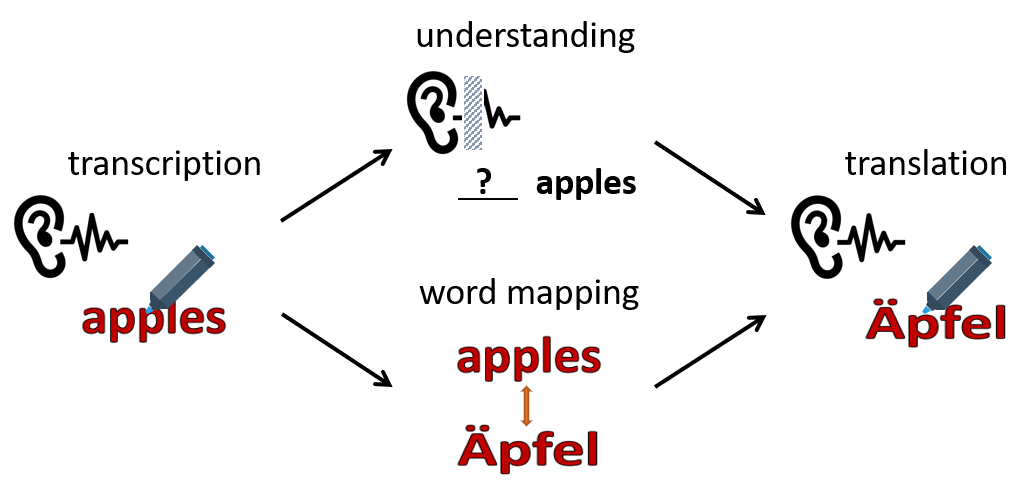}
    \end{minipage}
    }
    \end{subfigure}
    \caption{Comparison between previous encoder pre-training method with our curriculum pre-training method.}
    \label{intro}
\end{figure}

Due to the alleviation of error propagation and lower latency, the end-to-end ST model has
 been a hot topic in recent years.
However, large paired data of source audios and target sentences is required to train such a model, which is not easy to satisfy for most language pairs.  To address this issue, previous works resort to pre-training technique \cite{DBLP:conf/icassp/BerardBKP18, DBLP:conf/naacl/BansalKLLG19}, where they leverage the available ASR and MT data to pre-train an ASR model and an MT model respectively, and then initialize the ST model with the ASR encoder and the MT decoder. This strategy can bring faster convergence and better results.

The end-to-end ST encoder has three inherent roles: transcribe the speech, extract the syntactic and semantic knowledge of the source sentence and then map it to a semantic space, based on which the decoder can generate the correct target sentence. This poses a heavy burden to the encoder, which can be alleviated by pre-training. However, we argue that the current pre-training method restricts the power of pre-trained representations. The encoder pre-trained on the ASR task mainly focuses on transcription, which learns the alignment between the acoustic feature with phonemes or words, and has no ability to capture linguistic knowledge or understand the semantics, which is essential for translation. 

In order to teach the model to understand the sentence and incorporate the required knowledge, extra courses should be taken before learning translation.  Motivated by this, we propose a curriculum pre-training method for end-to-end ST. As shown in Figure \ref{intro}, we first teach the model \textbf{transcription} through ASR task. After that, we design two tasks, named frame-based masked language model (FMLM) task and frame-based bilingual lexicon translation (FBLT) task, to enable the encoder to \textbf{understand} the meaning of a sentence and \textbf{map words} in different languages. Finally, we fine-tune the model on ST data to obtain the  \textbf{translation} ability.

For the FMLM task, we mask several segments of the input speech feature, each of which corresponds to a complete word. Then we let the encoder predict the masked word. This task aims to force the encoder to recognize the content of the utterance and understand the inner meaning of the sentence. In FBLT, for each speech segment that aligns with a complete word, whether or not it is masked, we ask the encoder to predict the corresponding target word. In this task, we give the model more explicit and strong cross-lingual training signals. Thus, the encoder has the ability to perform simple word translation and the burden on the ST decoder is greatly reduced.
Besides, we adopt a hierarchical manner where different layers are guided to perform different tasks (first 8 layers for ASR and FMLM pre-training, and another 4 layers for FBLT pre-training). This is mainly because the three pre-training tasks have different requirements for language understanding and different output spaces. The hierarchical pre-training method can make the division of labor more clear and separate the incorporation of source semantic knowledge and cross-lingual alignments.

We conduct experiments on the LibriSpeech En-Fr and IWSLT18 En-De speech translation tasks, demonstrating the effectiveness of our pre-training method. The contributions of our paper are as follows: (1) We propose a novel curriculum pre-training method with three courses: transcription, understanding and mapping, in order to force the encoder to have the ability to generate necessary features for the decoder. (2) We propose two new tasks to learn linguistic  features, FMLM and FBLT, which explicitly teach the encoder to do source language understanding and target language meaning mapping. (3) Experiments show that both the proposed courses are helpful for speech translation, and our proposed curriculum pre-training leads to significant improvements.

\section{Related Work}
\subsection{Speech Translation}
Early work on speech translation used a cascade of an ASR model and an MT model \cite{DBLP:conf/icassp/Ney99,DBLP:conf/interspeech/MatusovKN05,DBLP:conf/icassp/MathiasB06}, which makes the MT model access to ASR errors. Recent successes of end-to-end models in the MT field \cite{DBLP:journals/corr/BahdanauCB14,DBLP:conf/emnlp/LuongPM15,DBLP:conf/nips/VaswaniSPUJGKP17} and the ASR fields \cite{DBLP:conf/icassp/ChanJLV16,DBLP:conf/icassp/ChiuSWPNCKWRGJL18} inspired the research on end-to-end speech-to-text translation system, which avoids error propagation and high latency issues.

In this research line, \citet{DBLP:journals/corr/BerardPSB16} give the first proof of the potential for an end-to-end ST model. After that, pre-training, multitask learning, attention-passing and knowledge distillation have been applied to improve the ST performance \cite{DBLP:conf/emnlp/AnastasopoulosC16,DBLP:conf/naacl/DuongACBC16,DBLP:conf/icassp/BerardBKP18,DBLP:conf/interspeech/WeissCJWC17,DBLP:conf/interspeech/BansalKLLG18,DBLP:conf/naacl/BansalKLLG19,DBLP:journals/tacl/SperberNNW19,DBLP:journals/corr/abs-1904-08075,DBLP:conf/icassp/JiaJMWCCALW19}. However, none of them attempt to guide the encoder to learn linguistic knowledge explicitly.
Recently,  \citet{DBLP:journals/corr/abs-1909-07575} propose to stack an ASR encoder and an MT encoder as a new ST encoder, which incorporates acoustic and linguistic knowledge respectively. However, the  gap between these two encoders is hard to bridge by simply concatenating the encoders.  \citet{DBLP:conf/interspeech/KanoS017} propose structured-based curriculum learning for English-Japanese speech translation, where they use a new decoder to replace the ASR decoder and to learn the output from the MT decoder (fast track) or encoder (slow track). They formalize learning strategies from easier networks to more difficult network structures. In contrast, we focus on the curriculum learning in pre-training and increase the difficulty of pre-training tasks.

\begin{figure*}
    \centering
    \includegraphics[width=0.95\textwidth]{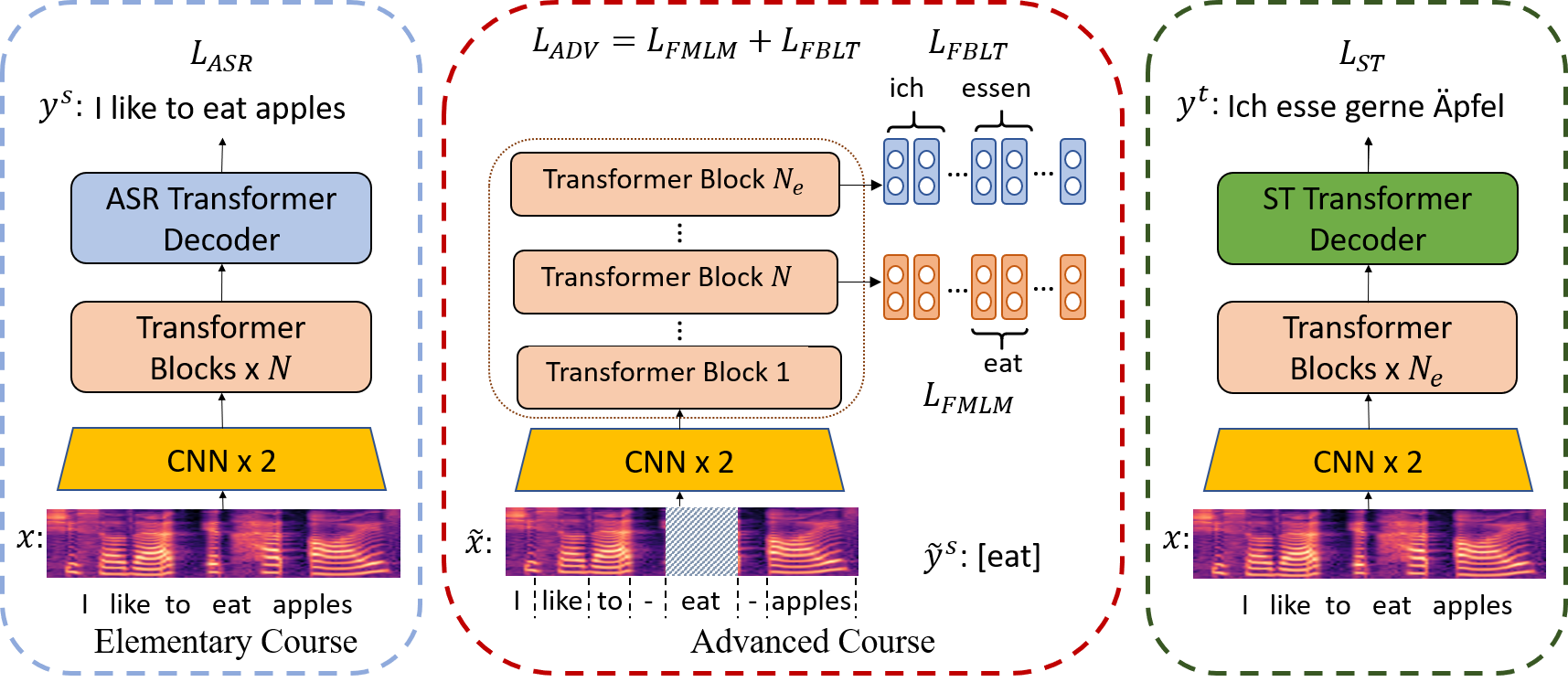}
    \caption{Proposed curriculum pre-training process. $\mathcal{L}_{FMLM}$ only predicts the mask word, while $\mathcal{L}_{FBLT}$ predicts all words in the target language.}
    \label{fig:model}
\end{figure*}

\subsection{Curriculum Learning}
Curriculum learning is a learning paradigm that starts from simple patterns and gradually increases to more complex patterns. This idea is inspired by the human learning process and is first applied in the context of machine learning by \citet{DBLP:conf/icml/BengioLCW09}. The study shows that this training approach results in better generalization and speeds up the convergence. Its effectiveness has been verified in multiple tasks, including shape recognition \cite{DBLP:conf/icml/BengioLCW09}, object classification \cite{DBLP:journals/tip/GongTMLKY16}, question answering \cite{DBLP:conf/icml/GravesBMMK17}, etc. However, most studies focus on how to control the difficulty of the training samples and organize the order of the learning data in the context of single-task learning. 

Our method differs from previous works in two ways: (1) We leverage the idea of curriculum learning for pre-training. (2) We do not train the model on the ST task directly with more and more difficult training examples or use more and more difficult structures. Instead, we design a series of tasks with increased difficulty to teach the encoder to incorporate diverse knowledge.

\section{Method}
\subsection{Overview}
The overview of our training process is shown in Figure \ref{fig:model}. It can be divided into three steps: First, we train the model towards the ASR objective $L_{ASR}$ to learn transcription. We note this as the elementary course. Next, we design two advanced courses (tasks) to teach the model understanding a sentence and mapping words in two languages, named Frame-based Masked Language Model (FMLM) task and Frame-based Bilingual Lexicon Translation (FBLT) task. In the FMLM task, we mask some speech segments and ask the encoder to predict the masked words. In the FBLT task, we ask the encoder to predict the target word for each speech segment which corresponds to a complete source word. In this stage, the encoder is updated by $L_{ADV}$. We adopt a hierarchical training manner where $N$ encoder blocks are used to perform ASR and FMLM tasks, since they both require outputs in source word space, and $N_e$ blocks are used in FBLT task. After the two-phases pre-training, the encoder is finally combined with a new decoder or a pre-trained MT decoder to perform the ST task towards $L_{ST}$.

\paragraph{Problem Formulation}
The speech translation corpus usually contains speech-transcription-translation triples, denoted as $\mathcal{S} =\{ (\bm{x}, \bm{y^{s}}, \bm{y^{t}})\}$. Specially, $\bm{x} = (x_1, \cdots, x_{T_x})$ is a sequence of acoustic features which are extracted from the speech signals. $\bm{y^{s}} = (y_1^{s}, \cdots, y_{T_s}^{s})$ and $\bm{y^{t}} = (y_1^{t}, \cdots, y_{T_t}^{t})$ represent the corresponding transcription in source language and the translation in target language respectively.
To pre-train the encoder, an extra ASR dataset $\mathcal{A} =\{ (\bm{x}, \bm{y^{s}})\}$ can be leveraged . Finally, the data for encoder pre-training is denoted as $\{ (\bm{x}, \bm{y^{s}})| (\bm{x}, \bm{y^{s}}) \in \mathcal{A} \vee (\bm{x}, \bm{y^{s}}, \bm{y^{t}}) \in \mathcal{S} \}$

After the encoder is pre-trained, we fine-tune the model using only $\mathcal{S}$, to enable it generate $\bm{y^t}$ from $\bm{x}$ directly. The model is updated using cross-entropy loss $  \mathcal{L}_{ST} = - \log P(\bm{y^t}|\bm{x})$.

\paragraph{Model Architecture}
In this work, we adopt the architecture of Transformer as in  \cite{DBLP:conf/asru/KaritaWWYZCHHIJ19}.  The  encoder is a stack of two $3 \times 3$ 2D CNN layers with stride 2 and $N_e$ Transformer encoder blocks. The CNN layers result in downsampling by a factor of 4. The decoder is a stack of $N_d$ Transformer decoder blocks.
\subsection{Elementary Course: Transcription}
In the elementary course, we train an end-to-end ASR model, which has the similar architecture as the ST model. The ASR encoder consists of $N$ blocks, and these blocks are used to initialize the bottom $ N$ blocks of the ST encoder.  For the ASR task, we follow  \citet{DBLP:conf/asru/KaritaWWYZCHHIJ19}, to employ a multi-task learning strategy, that is, both the E2E decoder and a CTC module predict the source sentence. Offline experiments indicate that the CTC objective is crucial for attentional encoder-decoder based ASR models. The final objective combines the CTC loss $\mathcal{L}_{ctc}$ and the cross-entropy loss $\mathcal{L}_{CE}$:
\begin{equation} \small
    \begin{aligned}
    \mathcal{L}_{ASR} & = \alpha \mathcal{L}_{CTC} + (1-\alpha) \mathcal{L}_{CE} \\
      & = -\alpha \log P_{ctc}(\bm{y^s}|\bm{x}) - (1-\alpha) \log P_{s2s}(\bm{y^s}|\bm{x})
    \end{aligned}
\end{equation}

In this work, we set $\alpha$ to 0.3. The CTC loss works on the encoder output and it pushes the encoder to learn frame-wise alignment between speech with words. The cross-entropy loss works on both the encoder and the ASR decoder.

\subsection{Advanced Courses: Understanding and Word Mapping}
With the ability of transcription, we further propose two new tasks for the advanced courses.
\subsubsection{Frame-based Masked Language Model}
The design of the Frame-based Masked Language Model task is inspired by the Masked Language Model (MLM) objective of BERT \cite{DBLP:conf/naacl/DevlinCLT19}, and semantic mask for ASR task \cite{DBLP:journals/corr/abs-1912-03010}. This task enables the encoder to understand the inner meaning of a segment of speech.

As shown in Figure \ref{fig:model},
we first perform force-alignment between the speech and the transcript sentence to determine where in time particular words occur in the speech segment. For each word ${y}^s_i$, we obtain its corresponding start position $s_i$ and the end position $e_i$ in the sequence $\bm{x}$ according to force alignment results.  At each training iteration, we randomly sample some percentage of the words in the $\bm{y^s}$ and denote the selected word set as $\bm{\tilde{y}^s}$.  Next, for each selected token $y_j^s$ in $\tilde{\bm{y}}^s$, we mask the corresponding speech piece $[{x}_{s_j}: {x}_{e_j}]$. The masked utterance is denoted as $\tilde{\bm{x}}$ and used as input to the encoder:
\begin{equation}
    \bm{h} = \text{Enc}(\tilde{\bm{x}})
\end{equation}

After that, for a masked piece $[{x}_{s_j}: {x}_{e_j}]$, we average the corresponding output hidden states $[h_{\lfloor \frac{s_j}{4} \rfloor}:h_{\lceil \frac{e_j}{4} \rceil}]$\footnote{The position indexs are divided by 4 due to downsampling.}, and compute the distribution probability over source words as shown in follows:
\begin{flalign}
\tilde{h}_j & = \text{mean}([h_{\lfloor \frac{s_j}{4} \rfloor}:h_{\lceil \frac{e_j}{4} \rceil}]) \\
p({y}^s_j|\tilde{\bm{x}}) & = \text{softmax}(\tilde{h}_j \cdot W)
\end{flalign}

In practice, the sentence is represented in BPE tokens and $W \in \mathcal{R}^{d_{model} \times |V_s|}$, where $|V_s|$ is the size of source vocabulary. In this way, a speech piece can be aligned with one or more tokens. We compute KL-Divergence loss as:
\begin{equation}
    \mathcal{L}_{FMLM} = - \sum_{{y}^s_j \in \tilde{\bm{y}}^s}\sum q({y}^s_j)\text{log} \frac{p({y}^s_j|\tilde{\bm{x}})}{q({y}^s_j)}
\end{equation}
$q({y}^s_i) \in \mathcal{R}^{|V_s|}$ is a distribution over all BPE tokens in source vocabulary $V_s$ and defined as:
\begin{equation}
q({y}^s_j)_{(pos)} = \left\{
             \begin{array}{ll}
             1/n_j, & V_s[pos] \in {y}^s_j \\
             0, &\mathrm{otherwise.}
             \end{array}
\right.
\label{Wc}
\end{equation}
where $pos$ represents the dimension index and $n_j$ is the total number of BPE tokens contained in word ${y}^s_j$.

 In this work, we use a mask ratio of 15\% following BERT and the masked speech piece is filled with the mean value of the whole utterance following \citet{DBLP:journals/corr/abs-1904-08779}. Because FMLM focuses on the understanding of source language, we computes its loss at the $N$-th layer of encoder (same with ASR loss), in the hope that the bottom $N$ layers are only concerned with source language.

\subsubsection{Frame-based Bilingual Lexicon Translation}
Aside from predicting masked source words, we go further to leverage cross-lingual information. Specifically, for each segment of speech features $[x_{s_i}:x_{e_i}]$ which aligned with a source word $y^s_i$, we assume we can obtain its target counterpart $\tilde{y}^t_i$. Similar to FMLM, we average the output hidden states from position $\lfloor \frac{s_i}{4} \rfloor$ to $\lceil \frac{e_i}{4} \rceil$, and then compute the distribution probability over target vocabulary. The alignment between speech segments and target words is a many-to-many correspondence, so there are cases where $\tilde{y}^t_i$ contains nothing or contains multiple foreign words. For the former case, we set the loss to zero, and for the latter case, we also compute KL-Divergence loss as:
\begin{equation}
    \mathcal{L}_{FBLT} = - \sum_{\tilde{y}^t_i} \sum q(\tilde{y}^t_i)\text{log} \frac{p(\tilde{y}^t_i|\tilde{\bm{x}})}{q(\tilde{y}^t_i)}
\end{equation} The definition of $q(\tilde{y}^t_i)$ is the length normalized distribution over all tokens appear in $\tilde{y}^t_i$. Note that the loss is computed on every speech segments, whether or not it is masked.

The only question remaining is how to obtain $\tilde{y}^t_i$ for each speech segment. Since there are two types of data for pre-training, $(\bm{x}, \bm{y^{s}}, \bm{y^{t}}) \in \mathcal{S}$ and  $(\bm{x}, \bm{y^{s}}) \in \mathcal{A} $, we use two methods to get the alignment:

$ \forall (\bm{x}, \bm{y^{s}}, \bm{y^{t}}) \in \mathcal{S}$, we simply run Moses\footnote{\url{http://www.statmt.org/moses}} scripts to establish word alignments. It begins from running of GIZA++\footnote{\url{https://github.com/moses-smt/giza-pp}} to get source-to-target and target-to-source alignments, and then runs a  heuristic grow-diag-final algorithm to get the final results, which means $\forall y_i^s \in  \bm{y^{s}}$, we choose one word from its translation sentence as the corresponding word $ \exists \tilde{y}^t_i \in \bm{y^t} \mbox{ s.t. } \tilde{y}^t_i \sim  \bm{y^{s}}  $.

Through the above alignment process, we can calculate a bilingual lexical translation table $\mathcal{T}$  with $\{ (\bm{y^{s}}, \bm{y^{t}}) |(\bm{x}, \bm{y^{s}}, \bm{y^{t}}) \in \mathcal{S}\}$, which estimates the translation probability between a source word $w^s_i$ and a target word $w^t_j$, denoted as $\mathcal{T} = {(w^s_i, w^t_j, p(w^s_i, w^t_j))}$. After that, $\forall (\bm{x}, \bm{y^{s}}) \in \mathcal{A} $, we compute a $\tilde{y}^t_i$ for each $y^s_i$ in $\bm{y^{s}}$ according to $\tilde{y}^t_i = \text{argmax}_{w^s_j} p(y^s_i, w^s_j)$.

We compute the $\mathcal{L}_{FBLT}$ at the top layer of the encoder, indicating that the top $N_e - N$ layers are duty on bilingual word mapping.  The final training objective in the advanced course combines FMLM and FBLT losses
\begin{equation}
    \mathcal{L}_{ADV} = \mathcal{L}_{FMLM} + \mathcal{L}_{FBLT}
\end{equation}

\section{Experiments}
\subsection{Data and Preprocess}\label{dataset}
We conduct experiments on two publicly available speech translation datasets: the LibriSpeech En-Fr Corpus \cite{DBLP:conf/lrec/KocabiyikogluBK18} and the IWSLT En-De Corpus \cite{niehues2018iwslt}.

\paragraph{LibriSpeech En-Fr:}
This corpus is a subset of the LibriSpeech ASR corpus \cite{DBLP:conf/icassp/PanayotovCPK15} and aligned with French e-books, which contains 236 hours of speech in total. Following previous works, we use the 100 hours clean training set and double the ST size by concatenating the aligned references with the provided Google Translate references, resulting in 90k training instances. We validate on the \textit{dev} set and report results on the \textit{test} set (2048 utterances).
\paragraph{IWSLT En-De:}
The corpus contains 271 hours of data, with English wave, English transcription, and German translation in each example. We follow \citet{DBLP:conf/asru/InagumaDKW19} to remove utterances of low alignment quality, resulting in 137k utterances. We sample 2k segments from the ST-TED corpus as dev set and \textit{tst2013} is used as the test set (993 utterances).

\paragraph{Data Preprocessing:}
We run ESPnet\footnote{\url{https://github.com/espnet/espnet}} \cite{DBLP:conf/interspeech/WatanabeHKHNUSH18} recipes to perform data pre-processing. For both tasks, our acoustic features are 80-dimensional log-Mel filterbanks stacked with 3-dimensional pitch features extracted with a step size of 10ms and window size of 25ms. The features are normalized by the mean and the standard deviation for each training set. Utterances of more than 3000 frames are discarded. We perform speed perturbation with factors 0.9 and 1.1. The alignment results between speech and transcriptions are obtained by Montreal Forced Aligner \cite{DBLP:conf/interspeech/McAuliffeSM0S17}.

For  references pre-processing, we tokenize and lowercase all the text with the Moses scripts. For pre-training tasks, the vocabulary is generated using sentencepiece \cite{DBLP:conf/emnlp/KudoR18} with a fixed size of 5k tokens for all languages, and the punctuation is removed. For ST task, we normalize the punctuation using Moses and use the character-level vocabulary due to its better performance \cite{DBLP:conf/icassp/BerardBKP18}.
Since there is no human-annotated segmentation provided in the IWSLT \textit{tst2013}, we use two methods to segment the audios: 1) Following ESPnet, we segment each audio with the LIUM SpkDiarization tool \cite{meignier2010lium}. For evaluation, the hypotheses and references are aligned using the MWER method with RWTH toolkit \cite{DBLP:conf/iwslt/BenderZMN04}. 2) We perform sentence-level force-alignment between audio and transcription using aeneas\footnote{\url{https://www.readbeyond.it/aeneas}} tool and segment the audio according to alignment results.

\subsection{Baselines}
Experiments are conducted in two settings: \textbf{base setting} and \textbf{expanded setting}. In base setting, only the corpus described in Section \ref{dataset} is used for each task. In the expanded setting, additional ASR and/or MT data can be used. All results are reported on case-insensitive BLEU with the multi-bleu.perl script unless noted.

\subsubsection{End-to-End ST Baselines}
We mainly compare our method with the conventional encoder pre-training method which uses only the ASR task to pre-train the encoder. Besides, we also compare with the results of the other works in the literature by copying their numbers.

\paragraph{LibriSpeech:} In the context of \textbf{base setting},  \citet{DBLP:conf/icassp/BerardBKP18} and ESPnet have reported results on a LSTM-based ST model with pre-training and/or multi-task learning strategy. \citet{DBLP:journals/corr/abs-1904-08075} use a Transformer ST model and knowledge distillation method. \citet{DBLP:journals/corr/abs-1909-07575} stack an ASR encoder and an MT encoder for final ST task, named as TCEN. Regarding the \textbf{expanded setting}, \citet{DBLP:journals/corr/abs-1911-08876} apply the SpecAugment on ST task. They use the total 236h of speech for ASR pre-training. \citet{DBLP:conf/asru/InagumaDKW19} combine three ST datasets of 472h training data \footnote{LibriSpeech En-Fr, IWSLT En-De and Fisher-CallHome Es-En} to train a multilingual ST model. In our work, we use the LibriSpeech ASR corpus as additional pre-training data, including 960h of speech. As the $dev$ and $test$ set of LibriSpeech ST task are extracted from the 960h corpus, we exclude all training utterances with the same speaker that appear in $dev$ or $test$ sets .

\paragraph{IWSLT:} Since previous works use different segmentation methods and BLEU-score scripts, it is unfair to copy their numbers. In our work, we choose the ESPnet results as {base setting} baseline, the multilingual model and TCEN-LSTM model as expanded baselines. \citet{DBLP:conf/asru/InagumaDKW19} use the same multilingual model as described in LibriSpeech baselines. And \citet{DBLP:journals/corr/abs-1909-07575} use an additional 272h TEDLIUM2\cite{DBLP:conf/lrec/RousseauDE14} ASR corpus and 41M parallel data from WMT18 and WIT3\footnote{\url{https://wit3.fbk.eu/mt.php?release=2017-01-trnted}}. All of them use ESPnet code, LIUM segmentaion method and multi-bleu.perl script. We follow \citet{DBLP:journals/corr/abs-1909-07575} to use another 272h ASR data for encoder pre-training and a subset of WMT18\footnote{Europarl v7, Common Crawl, News Comentary v13 and Rapid corpus of EU press releases.} for decoder pre-training.  We use the same processing method for MT data, resulting in 4M parallel sentences in total. We also re-implement the CL-fast track of \citet{DBLP:conf/interspeech/KanoS017} using our model architecture and data as another baseline.

\subsubsection{Cacased Baselines}
For LibriSpeech ST task, we use results of \citet{DBLP:conf/icassp/BerardBKP18}, \citet{DBLP:conf/asru/InagumaDKW19} and \citet{DBLP:journals/corr/abs-1904-08075} as base cascaded baselines. The first two use LSTM models for ASR and MT. While the last work trains Transformer ASR and MT models. We build an expanded cascaded system with the pre-trained Transformer ASR model and a LSTM MT model with the default setting in ESPnet recipe. For IWSLT ST task, we use \citet{DBLP:conf/asru/InagumaDKW19} as base cascaded baseline, which is based on LSTM architecture. And we implement a Transformer-based baseline using our pre-trained ASR and MT models in the expanded setting.
\begin{table*}[t] \small
\centering
\begin{tabular}{l|cc|c}
\hline
Method   & Enc pre-train & Dec pre-train   & BLEU  \\ \hline
MT\cite{DBLP:conf/icassp/BerardBKP18}*   &  -   &  -  & 19.3  \\
MT\cite{DBLP:conf/asru/InagumaDKW19}   &  -  & -   &  18.3 \\ \hline
\textbf{base setting}  &    &  &  \\
 LSTM ST \cite{DBLP:conf/icassp/BerardBKP18}*  &    &     & 12.9  \\
  \,\,\,\,+pre-train+multitask \cite{DBLP:conf/icassp/BerardBKP18}*  & \checkmark    &  \checkmark     & 13.4  \\
 LSTM ST+pre-train (ESPnet) & \checkmark & \checkmark & 16.68 \\
 Transformer+pre-train \cite{DBLP:journals/corr/abs-1904-08075}  & \checkmark  &  \checkmark  & 14.30 \\
 \,\,\,\, +knowledge distillation\cite{DBLP:journals/corr/abs-1904-08075} &  &  & 17.02\\
 TCEN-LSTM \cite{DBLP:journals/corr/abs-1909-07575} & \checkmark  &  \checkmark & 17.05 \\
 Transformer+ASR pre-train   & \checkmark  &  & 15.97 \\
 Transformer+curriculum pre-train  & \checkmark   &    & \bf{17.66}  \\ \hline
 \textbf{expanded setting}  &  &  &  \\
 LSTM+pre-train+SpecAugment\cite{DBLP:journals/corr/abs-1911-08876} & \checkmark (236h)  & \checkmark & 17.0 \\
 Multilingual ST+pre-train \cite{DBLP:conf/asru/InagumaDKW19} & \checkmark (472h) &  & {17.6} \\
 Transformer+ASR pre-train   & \checkmark (960h)   &    & 16.90 \\
 Transformer+curriculum pre-train &  \checkmark (960h) &  & \bf{18.01} \\ \hline
\end{tabular}
\caption{Comparison on LibriSpeech En-Fr test set. The size of ASR data for base setting is 100h unless labeled. Since inputs of the MT models are ground-truth text, the results of MT models can be seen as the upper-bound of ST models. *: Unknown BLEU score script.}
\label{Librispeech literature}
\end{table*}

\subsection{Implementation Details}
 All our models are implemented based on ESPnet.
 We set the model dimension $d_{model}$ to 256, the head number $H$ to 4, the feed forward layer size $d_{ff}$ to 2048. For LibriSpeech expanded setting, $d_{model}=512$ and $H=8$. For all the ST models, we set the number of encoder blocks $N_e = 12$ and the number of decoder blocks $N_d = 6$. Unless noted, we use $N=8$ encoder blocks to perform the ASR and the FMLM pre-training tasks. For MT model used in IWSLT expanded setting, we use the Transformer architecture in \citet{DBLP:conf/nips/VaswaniSPUJGKP17} with $N_e = 6, N_d = 6, H = 4, d_{model} = 256$.

 We train the model with 4 Tesla P40 GPUs and batch size is set to 64 per GPU. The pre-training takes 50 and 20 epochs for each phase and the final ST task takes another 50 epochs (a total of 120 epochs). We use the Adam optimizer with warmup steps 25000 in each phase. The learning rate decays proportionally to the inverse square root of the step number after 25000 steps. We save checkpoints every epoch and average the last 5 checkpoints as the final model. To avoid over-fitting, SpecAugment strategy \cite{DBLP:journals/corr/abs-1904-08779} is used in ASR pre-training with frequency masking (F = 30, mF = 2) and time masking (T = 40, mT=2). The decoding process uses a beam size of 10 and a length penalty of 0.2.

\subsection{Experimental Results}
\subsubsection{Comparison with End-to-End Baselines}
\textbf{LibriSpeech En-Fr:} The results on LibriSpeech En-Fr test set are listed in Table \ref{Librispeech literature}. In base setting, our method improves the ``Transformer+ASR pre-train" baseline by 1.7 BLEU and beats all the previous works, even though we do not pre-train the decoder. It indicates that through a well-designed learning process, the encoder has a strong potential to incorporate large amount of knowledge. Our method beats a  knowledge distillation baseline, where an MT model is utilized to teach the ST model. The reason, we believe, is that our method gives the model more training signals and makes it easier to learn. We also outperform a TCEN baseline which includes two encoders.  Compared to them, our method is more flexible and incorporates all information into a single encoder, which avoids the representation gap between the two encoders.

As the ASR data size increases, the model performs better. In the expanded setting, we find the FBLT task performs poorly compared with the base setting. This is because the target word prediction task is dictionary-supervised in expanded setting rather than reference-supervised as in base setting. However, our method still outperforms the simple pre-training method by a large margin. Besides, it is surprising to find that the end-to-end ST model is approaching the performance of an MT model, which is the upper bound of the ST model since it accepts golden source sentence without any ASR errors. This further verifies the effectiveness of our method.

\begin{table*}[t] \small
    \centering
    \begin{tabular}{l|cc|cc} \hline
    \multirow{2}{*}{Method}  &  {Enc pre-train} & Dec pre-train & \multicolumn{2}{c}{segment method} \\
            & (speech data)    & (text data)        &     LIUM & aeneas \\ \hline
    \textbf{base setting}  &   &   &   & \\
    ESPnet   &   &  & 12.50  &  - \\
    \,\,\,\,+enc pre-train & \checkmark &   & 13.12  & - \\
    \,\,\,\,+enc dec pre-train & \checkmark & \checkmark & 13.54 & -  \\
    Transformer+ASR pre-train  &  \checkmark  &  & 15.35 & 17.10 \\
    Transformer+curriculum pre-train         &  \checkmark    &   & \bf{16.27} & \bf{19.29}   \\ \hline
    \textbf{expanded setting}  & & &  & \\
    Multilingual ST+pre-train\cite{DBLP:conf/asru/InagumaDKW19}  & \checkmark(472h)  &   & 14.6  & - \\
    TCEN-LSTM \cite{DBLP:journals/corr/abs-1909-07575}   & \checkmark(479h) & \checkmark(40M) & 17.65  &  - \\
    CL-fast\cite{DBLP:conf/interspeech/KanoS017}(re-implemented)         & \checkmark(479h)  &   & 14.33  & 16.23 \\
    Transformer+curriculum pre-train+dec pre-train  & \checkmark(479h)   &  \checkmark(4M)   & \bf{18.15}  &  \bf{20.35}  \\ \hline
    \end{tabular}
    \caption{ST results on IWSLT En-De tst2013 set.}
    \label{IWSLT Results}
\end{table*}

\noindent \textbf{IWSLT En-De:}
The results on IWSLT tst2013 are listed in Table \ref{IWSLT Results}, showing a similar trend as in LibriSpeech dataset. We find that the segmentation methods have a big influence on the final results. In the base setting, our method can improve the ASR pre-training baseline by 0.9 to 2.2 BLEU scores, depending on the segmentation methods. In the expanded setting, we find when combined with decoder pre-train, the performance is further improved and beats other expanded baselines.

\subsubsection{Comparison with Cascaded Baselines}
Table \ref{cascaded} shows comparison with cascaded ST systems. For the base setting of two tasks, our end-to-end model can achieve comparable or better results with cascaded methods. This shows the end-to-end model has powerful learning capabilities and combines the functions of two models. In the LibriSpeech expanded setting, when more ASR data is available, we also obtain a competitive performance.  This indicates our method can make a good use of ASR corpus and learn valuable linguistic knowledge other than simple acoustic information. However, when additional MT data is used, there is still a gap between the end-to-end method and the cascaded method. How to utilize bilingual parallel sentences to improve the E2E ST model is worth further studying.

\begin{table}[] \small
\centering
\begin{tabular}{l|c}
\hline
{Method}   & BLEU  \\ \hline
\textbf{LibriSpeech base setting}  &  \\
LSTM ASR+ MT\cite{DBLP:conf/icassp/BerardBKP18}    & 14.6  \\
LSTM ASR+ MT\cite{DBLP:conf/asru/InagumaDKW19}  & 15.8  \\
Transformer ASR + MT\cite{DBLP:journals/corr/abs-1904-08075} & 17.85\\
Ours E2E Transformer ST & 17.66  \\ \hline
\textbf{LibriSpeech expanded setting} \\
Transformer ASR+LSTM MT* & 18.05 \\
Ours E2E Transformer ST  & 18.01 \\
 \hline
\hline
\textbf{IWSLT base setting} &  \\
LSTM ASR+ MT\cite{DBLP:conf/asru/InagumaDKW19}  & 14.0  \\
Ours E2E Transformer ST  &  16.27    \\ \hline
\textbf{IWSLT expanded setting}  & \\
Transformer ASR+Transformer ST & 22.16 \\
Ours E2E Transformer ST & 18.15   \\
 \hline
\end{tabular}
\caption{Comparison with cascaded ST. *:we find the LSTM model outperforms Transformer model in our setting since the training data is scarce.}
\label{cascaded}
\end{table}
\subsection{Analysis and Discussion}
\textbf{Ablation Study}
To better understand the contribution of each component, we perform an ablation study on LibriSpeech expanded setting. The results are shown in Table \ref{ablation}.  On the one hand, we show that both of our proposed pre-training tasks are beneficial: In ``-FMLM task" and ``-FBLT task"\footnote{we use 12-layer encoder for ASR and FMLM pre-training for a fair comparison.}, we perform single-task pre-training for advanced course.  The performance drops when we remove either one of them. On the other hand, we show the two-phases pre-training paradigm is necessary: The ``- phase 2" experiment degenerates to the simple ASR pre-training baseline. In ``-phase 1" setting, we find that without the ASR pre-training, the training accuracy on FMLM task and FBLT task drops a lot, which further affects the ST performance. This means the ASR task is necessary for both the advanced courses and ST. In ``Multi3" setting, we pre-train the model on ASR, FMLM and FBLT tasks in one phase. In this setting, we observe multi-task learning also decrease individual task performances (ASR, FMLM and FBLT) compared to curriculum learning. One reasonable expanation is that it is hard to train on the FMLM and FBLT tasks which takes masked input from randomly initialized parameters, which also leads to performance degradation on the ST task.

\begin{table}[]  \small
    \centering
    \begin{tabular}{l|c}
    \hline
        Method & BLEU \\ \hline
       Our method  & \bf{18.01} \\
        \,\,\,\,-FMLM task  &  17.62 \\
      \,\,\,\,-FBLT task  &  17.65 \\
      \,\,\,\, -phase 2  &  16.90  \\
     \,\,\,\, -phase 1  &  14.26 \\
      Multi3  &   14.82 \\ \hline

    \end{tabular}
    \caption{Ablation study on LibriSpeech expanded setting. `-' indicates removing the task or phase from our method.}
    \vspace{-2mm}
    \label{ablation}

\end{table}

\noindent \textbf{Hyper-parameter $\mathbf{N}$}
During pre-training, which layer conducts  ASR pre-training and FMLM loss is an important  hyper-parameter. We conduct experiments on LibriSpeech base setting to explore the influence of different choices. We keep $N_e = 12$ unchanged and always use the top layer to perform the FBLT task. Then we alter the hyperparameter $N$. We find if $N=6$, the model finds it difficult to converge during ST training. That may be because the distance between the decoder and the bottom 6 encoder layers is too far so that the valuable source linguistic knowledge can not be well utilized. Moreover, the model performs undesirable if the choice is 10 or 12, which results in 16.47 and 16.14 BLEU score respectively, since the number of blocks for FBLT task is not enough. The model achieves the best performance when we choose $N=8$. Thus, we use this strategy in our main experiments.

\noindent \textbf{Unlabeled Speech Data}
In this work, we also explore how to utilize the unlabeled speech data in pre-training, but only get negative results. We conduct exploratory experiments on the LibriSpeech ST task. Assume that the $(\bm{x}, \bm{y^s})$ from 100h ST corpus as labeled pre-training data and $(\bm{x})$ from 960h LibriSpeech ASR corpus as unlabeled data. Following \citet{DBLP:journals/corr/abs-1910-09932}, we design an unsupervised pre-training task for elementary course, in which we randomly mask 15\% of fbank features and let the bottom 4 encoder layers predict the masked part. We compute the L1 loss between the prediction and groundtruth filterbanks. However, we find that this method is not helpful for the final ST task, which results in 16.85 BLEU score, lower than our base setting model (without extra data pre-training). It is still an open question about how to use unlabeled speech data.

\section{Conclusion and Future Work}
This paper investigates the end-to-end method for ST. We propose a curriculum pre-training method, consisting of an elementary course with an ASR loss, and two advanced courses with a frame-based masked language model loss and a bilingual lexicon translation loss, in order to teach the model syntactic and semantic knowledge in the pre-training stage. Empirical studies have demonstrated that our model significantly outperforms baselines. In the future, we will explore how to leverage unlabeled speech data and large bilingual text data to further improve the performance. Besides, we expect the idea of curriculum pre-training can be adopted on other NLP tasks.

\section*{Acknowledgements}
This work was supported in part by the National Natural Science Foundation of China under Grant No.U1636116 and the Ministry of education of Humanities and Social Science project under grant 16YJC790123.

\bibliography{acl2020}
\bibliographystyle{acl_natbib}
\end{document}